\documentclass[lettersize,journal]{IEEEtran}
\usepackage{amsmath,amsfonts}
\usepackage{algorithmic}
\usepackage{algorithm}
\usepackage{array}
\usepackage[caption=false,font=normalsize,labelfont=sf,textfont=sf]{subfig}
\usepackage{textcomp}
\usepackage{stfloats}
\usepackage{url}
\usepackage{verbatim}
\usepackage{graphicx}
\usepackage{cite}
\usepackage{bm}
\usepackage{xcolor}
\usepackage{bbm} 

\definecolor{skyblue}{RGB}{224, 255, 255}
\definecolor{citeblue}{RGB}{54, 158, 252}

\usepackage{amsmath}

\usepackage[T1]{fontenc}
\usepackage{booktabs} 
\usepackage{colortbl}
\usepackage{float}
\usepackage{threeparttable}
\usepackage{tabularx}
\usepackage{catchfilebetweentags}
\usepackage{xr}
\usepackage{makecell}
\usepackage{enumitem}

\usepackage{xspace}
\newcommand{\method}{\textsc{CRA}\xspace}

\begin{document}


\title{Contextual Representation Anchor Network for Mitigating Selection Bias in Few-Shot Drug Discovery}


\author{IEEE Publication Technology,~\IEEEmembership{Staff,~IEEE,}
\author{Ruifeng Li, Wei Liu, Xiangxin Zhou, Mingqian Li, Qiang Zhang,
 Hongyang Chen, ~\IEEEmembership{Senior member,~IEEE,}
  Xuemin Lin, ~\IEEEmembership{Fellow,~IEEE}}
\IEEEcompsocitemizethanks{
    \IEEEcompsocthanksitem Correspond to Hongyang Chen.
    \IEEEcompsocthanksitem Ruifeng Li is with the College of computer science and technology, Zhejiang University, Hanghzou, China. Email: lirf@zju.edu.cn.
    \IEEEcompsocthanksitem Wei Liu is with Department of Computer Science and Engineering, Shanghai Jiao Tong University, Shanghai, China. Email: captain.130@sjtu.edu.cn.
    \IEEEcompsocthanksitem Xiangxin Zhou is with School of Artificial Intelligence, University of Chinese Academy of Sciences and New Laboratory of Pattern Recognition (NLPR), State Key Laboratory of Multimodal Artificial Intelligence Systems (MAIS), Institute of Automation, Chinese Academy of Sciences (CASIA), Beijing, China. Email: zhouxiangxin1998@gmail.com.
    \IEEEcompsocthanksitem Mingqian Li is with the Research Center for Data Hub and Security, Zhejiang Lab, Hangzhou, China. Email: mingqian.li@zhejianglab.com.
    \IEEEcompsocthanksitem Qiang Zhang is with the College of Computer Science and Technology, Zhejiang University, Hangzhou, China, and also with ZJU-Hangzhou Global Scientific and Technological Innovation Center, Hangzhou, China. Email: qiang.zhang.cs@zju.edu.cn.
    \IEEEcompsocthanksitem Hongyang Chen is with the Research Center for Data Hub and Security, Zhejiang Lab, Hangzhou, China. Email: hongyang@zhejianglab.com.
    \IEEEcompsocthanksitem Xuemin Lin is with Antai College of Economics and Management, Shanghai Jiao Tong University, Shanghai, China. Email: xuemin.lin@sjtu.edu.cn.
    \IEEEcompsocthanksit
    \IEEEcompsocthanksitem Ruifeng Li and Wei Liu contributed equally to this work. \protect\\
}

\thanks{This paper was produced by the IEEE Publication Technology Group. They are in Piscataway, NJ.}
\thanks{Manuscript received April 19, 2021; revised August 16, 2021.}}

\markboth{Preprint}%
{Shell \MakeLowercase{\textit{et al.}}: A Sample Article Using IEEEtran.cls for IEEE Journals}


\maketitle

\begin{abstract}
In the drug discovery process, the low success rate of drug candidate screening often leads to insufficient labeled data, causing the few-shot learning problem in molecular property prediction. Existing methods for few-shot molecular property prediction overlook the sample selection bias,
which arises from non-random sample selection in chemical experiments. This bias in data representativeness leads to suboptimal performance.
To overcome this challenge, 
we present a novel method named 
\underline{c}ontextual \underline{r}epresentation 
\underline{a}nchor
network (\textbf{CRA}), 
where an anchor refers to a cluster center of the representations of molecules and serves as a bridge to transfer enriched contextual knowledge into molecular representations and enhance their expressiveness. CRA introduces a dual-augmentation mechanism that includes context augmentation, which dynamically retrieves analogous unlabeled molecules and captures their task-specific contextual knowledge to enhance the anchors, and anchor augmentation, which leverages the anchors to augment the molecular representations.


We evaluate our approach using the MoleculeNet and FS-Mol benchmarks, as well as through domain transfer experiments. The outcomes indicate that CRA surpasses current state-of-the-art methods by 2.60\% in AUC and 3.28\% in $\Delta$AUC-PR metrics, showcasing its exceptional generalization abilities.

\end{abstract}


\section{Introduction}

Accelerating drug discovery is crucial for effective disease management and public health~\cite{moffat2017opportunities, atanasov2021natural, thomford2018natural}.
Artificial intelligence (AI) has revolutionized this field by developing AI-driven drug discovery (AIDD) \cite{vamathevan2019applications, lavecchia2015machine, mak2023artificial}, which uses deep learning to streamline traditional processes. This significantly reduces the time and cost of predicting and designing new drug molecules \cite{tetko2001prediction, zhou2020artificial}. 
As an integral component of AIDD, Quantitative Structure-Activity/Property Relationship (QSAR/QSPR) models establish relationships between molecular structures and their biological activities or properties, enhancing the efficiency of drug discovery \cite{muratov2020qsar, kiralj2009basic, gramatica2013development, yee2012current}. 
Despite their success with large datasets \cite{fang2023knowledge, wang2022molecular, fang2022geometry, zhang2021motif}, the persistent scarcity of training data—stemming from the costly and time-consuming nature of chemical experiments with inherently low success rates—poses a significant challenge \cite{rajpurkar2022ai,anthony2020metallodrugs}. 
To overcome this, few-shot learning has emerged as a promising strategy, enabling enhanced molecular property prediction in data-limited scenarios \cite{altae2017low}.


Few-shot learning enables models to adapt and perform well on new tasks using only a few training examples. Recently, several methods, such as IterRefLSTM \cite{altae2017low}, Meta-MGNN \cite{guo2021few}, ADKF-IFT \cite{chen2023metalearning}, and PAR \cite{wang2021propertyaware}, have been proposed to address this challenge by leveraging metric-based methods or meta-learning strategies to generate molecular representations that can effectively generalize from a limited number of support set samples. These methods heavily rely on the support set to learn task-specific representations and establish decision boundaries between different molecular categories.

\begin{figure}
\centering\includegraphics[width=0.4\textwidth]{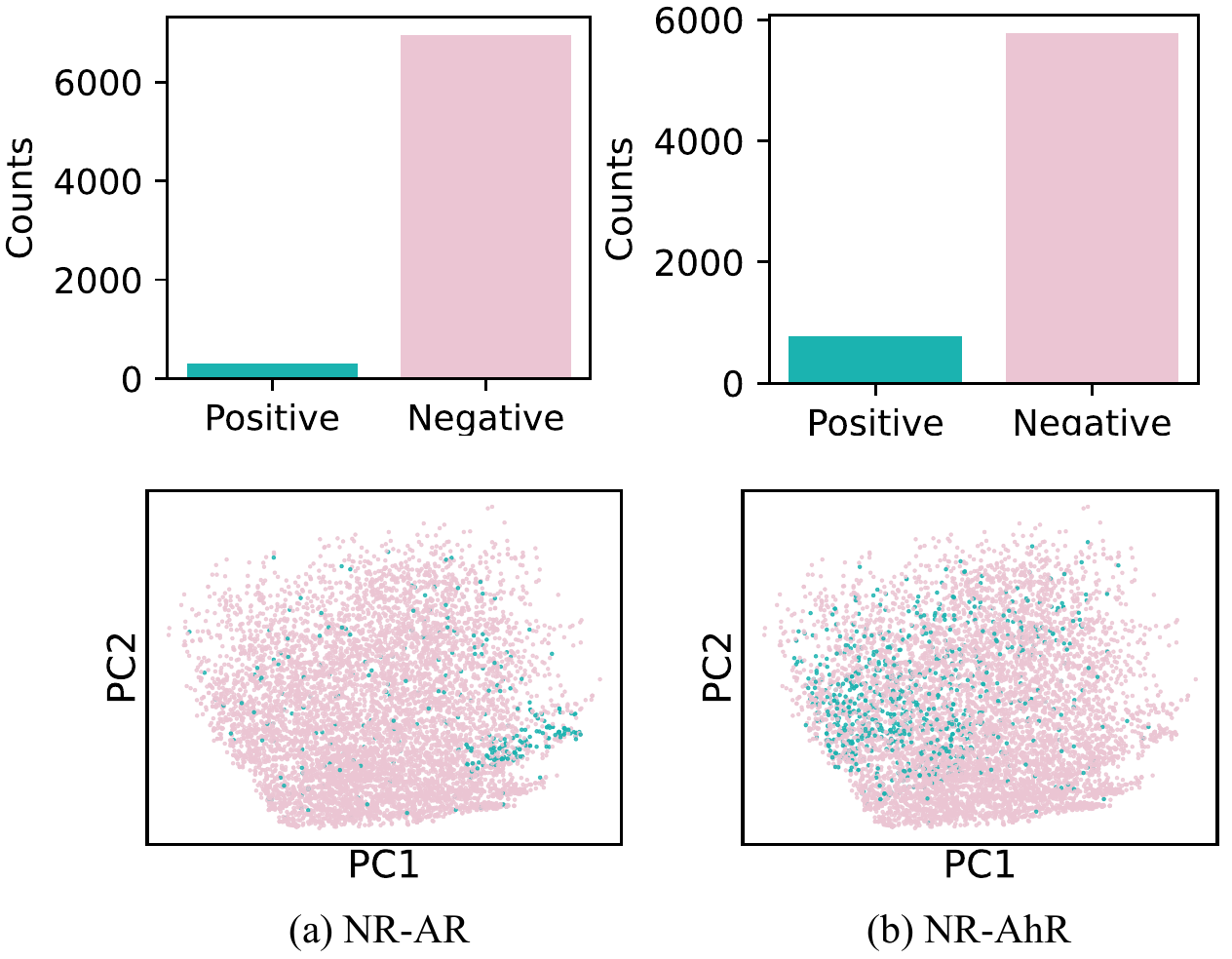} 
\caption{The sample selection bias in the NR-AR and NR-AhR tasks from the Tox21 dataset. The top row shows the distribution of positive and negative samples, highlighting the imbalance in data sample. The bottom row displays 2D PCA projections of the molecular MACCS fingerprints, illustrating the clustering patterns influenced by sample selection bias.}
 \label{fig:cpp_bias}
\end{figure}

However, the heavy reliance on small support sets makes these methods susceptible to sample selection bias. The bias arises when the samples in the support set fail to adequately represent the overall distribution of the molecular space. In drug discovery, samples often come from non-random chemical experiments and lack broad representativeness \cite{ellenberg1994selection, fan2007sample}. 
These samples typically cover only a subset of the entire chemical space, 
 causing the risk of selection bias  \cite{hernan2004structural}. 
Additionally, the significant domain gap between different tasks in drug discovery further exacerbates this challenge, as shown in Figure~\ref{fig:cpp_bias}. 
Therefore, addressing sample selection bias is crucial for solving the few-shot learning problem in drug discovery.


\begin{figure*}
\centering\includegraphics[width=0.75\textwidth]{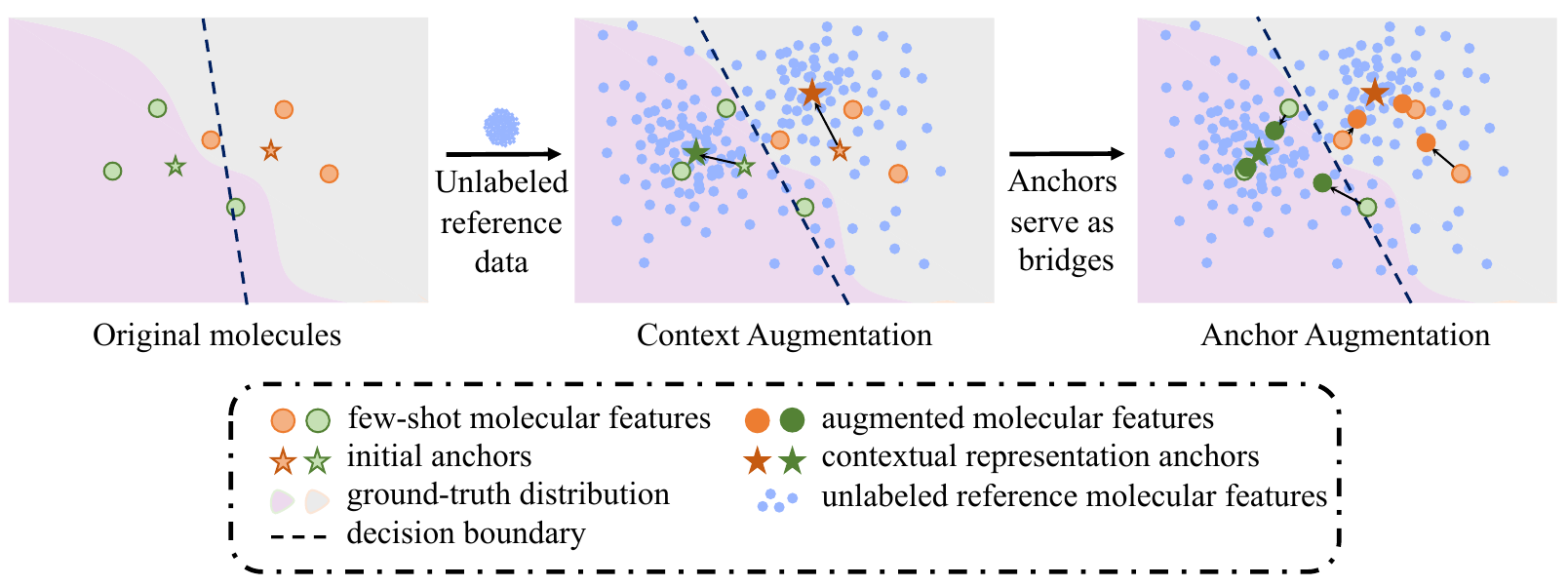}
    \caption{The augmentation of molecular embeddings and Contextual representation anchors in \method. The left part shows the initial molecular embeddings and anchors. The middle part shows the change of anchors after augment with reference data. The right part shows the change of molecular embeddings by using Contextual representation anchors for augmentation.}
    \label{fig:cpp_intro}
    \vspace{-15pt}
\end{figure*}


Inspired by contextual data augmentation strategies \cite{kobayashi2018contextualaugmentationdataaugmentation}, we utilize contextual molecules, i.e., abundant molecule samples without labels, to fill the knowledge gaps in the limited labeled samples. 
Contextual data augmentation effectively alleviates sample selection bias by incorporating diverse unlabeled samples to calibrate class-specific feature deviations, thereby preventing the model from overfitting to certain class distributions during training \cite{xu2022alleviating}. 
Although MHNfs \cite{schimunek2023contextenriched} utilize Modern Hopfield Networks \cite{ramsauer2020hopfield} for context-based molecular representation enhancement, their approach primarily focuses on improving individual molecular representations but lacks task-specific targeted enhancement. This may inadvertently amplify irrelevant features, increasing bias and leading to overfitting or poor generalization.

To address this issue, we propose the Contextual Representation Anchor Network (CRA), which introduces a dual-augmentation mechanism: first from context (i.e., unlabeled molecules) to anchors (i.e., class-level contextual representation anchors), and then from anchors to individual molecules (Figure~\ref{fig:cpp_intro}). 
The core function of the anchor is to connect labeled samples with unlabeled samples, bridging the information gap between different molecules. In this process, attention mechanisms serve as dynamic, task-sensitive tools to ensure that the anchors fully absorb contextual knowledge from unlabeled data and transfer this information precisely to the representation of labeled samples. This design not only enhances the consistency of intra-class features but also, through the bridging role of the anchors, improves the distinction between inter-class molecules, solving the limitations of traditional models in few-shot scenarios.




The main contributions of this work are summarized as follows:

\begin{itemize}[leftmargin=*,topsep=0pt,itemsep=0pt]
\item
 We propose 
the Contextual Representation Anchor Network (CRA), which, to the best of our knowledge, is the first method to address sample selection bias in few-shot molecular property prediction task.
CRA leverages class-level contextual representation anchors as bridges between labeled and unlabeled data with a dual-augmentation mechanism, effectively utilizing the contextual knowledge from unlabeled samples to enhance intra-class feature representation while preserving inter-class distinctiveness (Section~\ref{method}).

\item 
We present a context augmentation module that dynamically retrieves analogous molecules and captures task-specific contextual knowledge to augment the contextual representation anchors. By continuously integrating feedbacks from similar molecules, this module refines the positioning and structural relationships of the anchors within the feature space, ensuring that 
they accurately capture the nuanced co-occurrence patterns and dependencies between molecular properties
(Section~\ref{sss:context_calibrated_module}). 
\item 
We introduce an anchor augmentation module that utilizes contextual representation anchors as bridges to seamlessly transfer enriched contextual knowledge into molecular representations. This module refines intra-class co-occurrence structures and boosts the overall expressiveness of the molecular representations, resulting in more robust feature learning and improved representational quality
(Section~\ref{sss:prototype-prompt_attention}). 
\item
We evaluate our \method on the MoleculeNet and FS-Mol (Section~\ref{exp:fs_mol_benchmark}) benchmarks. The results show that it outperforms SOTA by an average of \textbf{2.60}\% (AUC) and \textbf{3.28}\% ($\Delta$AUC-PR) on the respective benchmarks.
Further experiments demonstrate the strong generalization ability of our model, with \textbf{3.70}\% (AUC) and \textbf{20.20}\% ($\Delta$AUC-PR) gains (Section~\ref{sub:domain_shift_experiment}).
\end{itemize}

\section{Related work}
\subsection{Few-shot Learning in Drug Discovery}
\label{para:few-shot learning in drug discovery}

Few-shot learning methods in drug discovery can be broadly categorized into transductive and inductive approaches. Transductive methods, such as embedding-based nearest neighbor techniques, map molecules into a chemical space to predict properties by analyzing inter-molecular relationships within a specific task. For example, IterRefLSTM \cite{altae2017low} employs iterative reasoning and LSTM networks, while PAR \cite{wang2021propertyaware} generates task-specific molecular representations to construct a relation graph for label propagation. Conversely, inductive methods, including optimization-based or fine-tuning approaches like GNN-MAML \cite{guo2021few} and ADKF-IFT \cite{chen2023metalearning}, leverage the meta-learning framework \cite{hospedales2021meta, patacchiola2020bayesian, finn2017model} to rapidly generalize to new molecules. 
While both methods have shown effectiveness, they often overlook the critical issue of sample selection bias. 
MHNfs \cite{schimunek2023contextenriched} enhance molecular representations using contextual molecules, but they struggle to effectively differentiate between intra-class and inter-class relationships.
Unlike MHNFs, 
our \method establishes dynamic connections between unlabeled data and intra-class feature representations through contextual representation anchors, enabling the model to more accurately enhance intra-class features and mitigate sample selection bias.

\subsection{Semi-supervised Few-shot Learning}
\label{para:semi-supervised few-shot learning}
Semi-supervised few-shot learning, a method that integrates a small amount of labeled data with a larger pool of unlabeled data, has gained significant attention in the field of machine learning. The commonly used strategy in this method is to train the labeled few-shot data together with the unlabeled data by using pseudo-labels \cite{huang2021pseudo, zhang2022sample, cui2023uncertainty}. For instance, Meta-SSFSL \cite{ren2018meta} incorporates soft k-means and unlabeled data to refine prototypes, surpassing the performance of ProtoNets \cite{snell2017prototypical}. PRWN \cite{ayyad2021semi} enhances few-shot learning models by implementing prototypical magnetization and global consistency to improve representation learning. Cluster-FSL \cite{ling2022semi} adopts Multi-Factor Clustering for generating pseudo-labels and integrates data augmentation techniques to achieve superior results. 
Semi-supervised learning is not only applicable to computer vision but also suitable for processing molecular graphs in cheminformatics, where the sparsity of molecular graphs requires additional chemical information to optimize predictions. 
In our method, we leverage the attention mechanism to implement the dual-augmentation mechanism, effectively strengthening the connections between the unlabeled data, anchors, and intra-class molecules, resulting in more precise and robust molecular representations.


\subsection{Sample Selection Bias}
\label{para:sample selection bias}

Sample selection bias, which leads to distorted results, occurs when samples fail to accurately represent the entire feature space, causing models that perform poorly on unseen data. This challenge is especially pronounced in few-shot learning due to the limited number of samples, making bias reduction a critical area of extensive research \cite{huang2006correcting, fan2022debiased}. Bias reduction techniques can be generally divided into normalization-based and calibration-based methods. For example, normalization-based approaches like Z-Score Normalization (ZN) \cite{fei2021z}, SimpleShot \cite{Wang2019SimpleShotRN}, and SEN \cite{nguyen2020sen} employ various normalization strategies to mitigate bias. On the other hand, calibration-based methods, such as the Distribution Calibration Method (DCM) \cite{tao2022powering} and Task-Calibrated Prototypical Representation (TCPR) \cite{xu2022alleviating}, develop specific strategies tailored to address this issue. Our method uniquely targets sample selection bias within drug discovery, specifically addressing the few-shot molecular property prediction challenge.Unlike these methods, we introduce contextual representation anchors as bridge which dynamically enhance inter-class molecular representations to mitigate sample bias.


\section{Preliminaries}
\label{section:preliminaries}
\subsection{Molecular Property Prediction}\label{subsection:MMP} Typically, deep learning models the function $f_{\bm{\theta}}$ to predict the molecular properties or activities $\hat{\mathbf{y}}$ based on the molecular features $\mathbf{x}$.  The model $f_{\bm{\theta}}$ consists of two main components: an encoder $f_e$ and a classifier $g$.
The encoder $f_e$ is capable of processing various molecular features $\mathbf{x}$, such as Simplified Molecular Input Line Entry System (SMILES) \cite{weininger1988smiles, winter2019learning}, descriptors \cite{bender2004similarity, unterthiner2014deep}, fingerprints \cite{glen2006circular, rogers2010extended}, and molecular graphs \cite{kearnes2016molecular, merk2018novo,  yang2019analyzing, jiang2021could}, then outputs high-dimensional molecular representations. The classifier $g$ usually takes the high-dimensional molecular representations as input to generate the predicted categories $\hat{y}$.

\subsection{Multi-Head Attention Mechanism}\label{subsection:MHA}
The multi-head attention mechanism with $H$ heads takes a query $\mathbf{Q}\in \mathbb{R}^{N_1 \times d_k}$, a key $\mathbf{K}\in \mathbb{R}^{N_2 \times d_k}$, and a value $\mathbf{V}\in \mathbb{R}^{N_2 \times d_k}$ as inputs, calculates $H$ attention results, and projects them into the output embedding space:
\begin{equation}
\begin{aligned}
    \label{eq:MHAM}
    \mathbf{Y} &= \operatorname{Multi-HeadAttention}(\mathbf{Q},\mathbf{K},\mathbf{V}) \\
    &=\sum_{i=1}^{H} \operatorname{softmax}\left(\dfrac{\mathbf{Q}\mathbf{W}_i^Q (\mathbf{K}\mathbf{W}_i^K)^{\top}}{\sqrt{d_k}}\right)\mathbf{V}\mathbf{W}_i^V \cdot \mathbf{W}_i^O, 
\end{aligned}
\end{equation}
    where $d_k$ is the dimension of the key, and all $\mathbf{W}_i^{*}$ $\in \mathbb{R}^{d_k \times d_k}$ are learnable parameter matrices. Here, we denote the multi-head self-attention with a residual connection as: 
\begin{equation}
\begin{aligned}
    \mathbf{X}^{\prime} = & \operatorname{R-MHA}(\mathbf{X}) \\= & \mathbf{X}+ \operatorname{Multi-Head Attention}(\mathbf{X},\mathbf{X},\mathbf{X}).    
\end{aligned}
\end{equation}

\section{Methodology}
\label{method}

\begin{figure*}
\centering\includegraphics[width=0.95\textwidth]{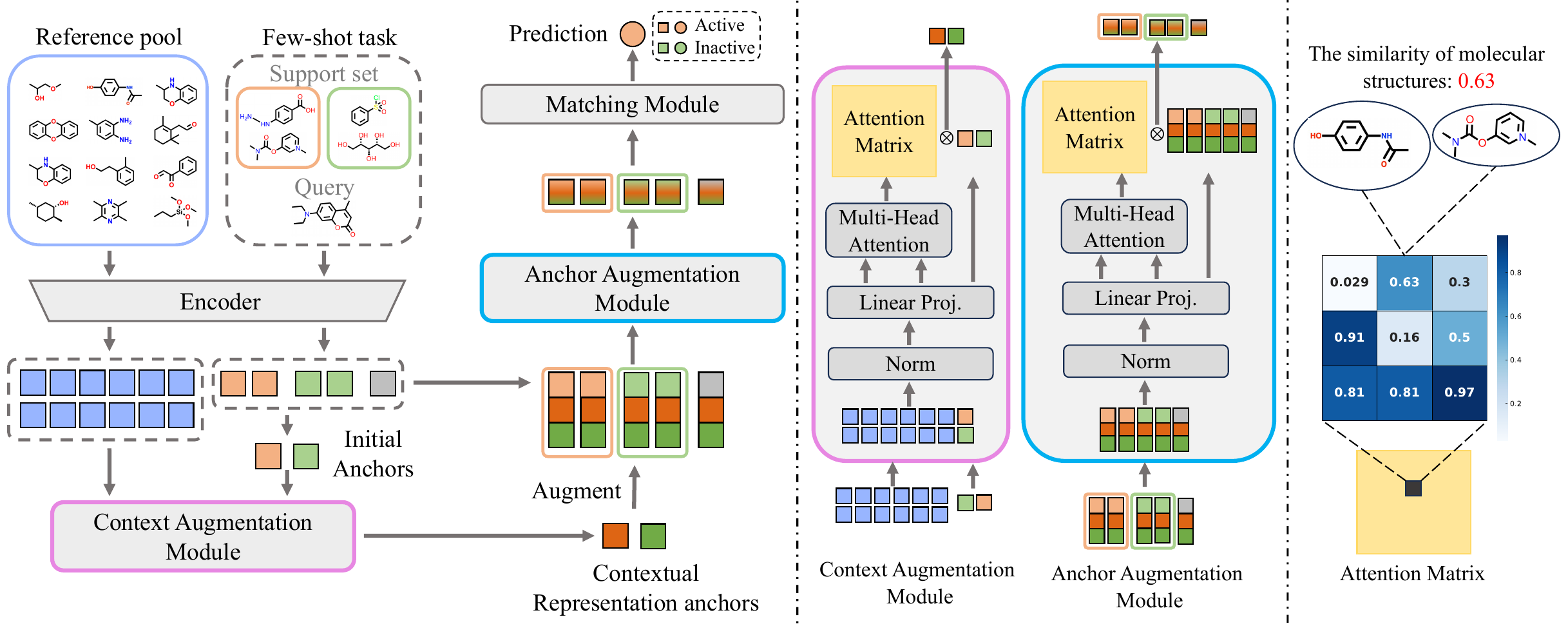}
    \caption{Overview of \method. The architecture comprises three main components: the context augmentation module (CAM), the anchor augmentation module (AAM), and the matching module (MM). The procedure begins with a shared encoder to obtain molecular embeddings and initial anchors. Next, The CAM augments these initial anchors with unlabeled reference molecules. Then, the augmented anchors are used to form augmented molecular embeddings, which are enhanced by the AAM. The MM leverages the similarities between the query samples and support samples to derive the final predictions. The attention matrix highlights the similarity between molecular structures (e.g., a similarity score of 0.63).}
    \label{fig:method_pipeline}
\end{figure*}

In this section, we present a detailed overview of our contextual representation anchor network (CRA), with its overall architecture illustrated in Figure~\ref{fig:method_pipeline}. 
First, we define the few-shot learning problem in the context of molecular property prediction and introduce the role of the attention mechanism.
Then, we describe the architecture of our CRA model, followed by an explanation of the CRA algorithm.
Finally, we provide a detailed introduction of the training and inference phases of our model.

\subsection{Problem Definition}
\label{subsection:problem_definition}
We consider the problem of few-shot molecular property prediction. A set of molecular property prediction tasks $\left\{\mathcal{T}_{\tau}\right\}_{\tau=1}^{N_{t}}$ are available for model training. Each task $\mathcal{T}_{\tau} =\left\{\left(\mathbf{x}_{\tau, i}, y_{\tau, i}\right)\right\}_{i=1}^{N_{\tau}}$,  where $\mathbf{x}_{\tau, i} \in \mathbb{R}^{d}$ represents the molecular features and $y_{\tau, i} \in \left\{-1, 1\right\}$ represents the molecular property or activity.
Let $N_{\tau,c}$ denote the number of molecules in class $c$ in $\mathcal{T}_{\tau}$, where $c \in \{-1, 1\}$.
The goal is to learn a generalized model $f_{\bm{\theta}}$ that can quickly adapt to new, previously unseen few-shot molecular property prediction tasks $\left\{\mathcal{T}_{\tau}\right\}_{\tau=N_{t}+1}^{N_{t}+N_{e}}$ in drug discovery. 
Specifically, a new few-shot learning task $\mathcal{T}_{\tau}$ comprises a support set $\mathcal{S}_{\tau}=\left\{\left(\mathbf{x}_{\tau, s}, y_{\tau, s}\right)\right\}_{s=1}^{N_{\tau}^{s}}$ to be learned with, and a query set  $\mathcal{Q}_{\tau}=\left\{\left(\mathbf{x}_{\tau, q}, y_{\tau, q}\right)\right\}_{q=1}^{N_{\tau}^{q}}$ to be classified. 
A large unlabeled molecular dataset $\mathcal{B}=\left\{\mathbf{x}_ b\right\}_{b=1}^{N^b}$ can be used as reference since it is easily accessible and has rich structural information. 

\subsection{Architecture}
\label{subsection:CRA_network}

In this work, we propose a simple yet effective method called the Contextual Representation Anchor Network (CRA). This method employs a dual-augmentation mechanism to dynamically enhance the intra-class co-occurrence structures of molecules, effectively mitigating bias.
As illustrated in Figure \ref{fig:method_pipeline}, \method consists of three key components: the context augmentation module (CAM), the anchor augmentation module (AAM), and the matching module (MM). 
First,  a shared encoder is used to obtain molecular embeddings for the support, query, and reference sets. Initial class-level anchors are then derived from the support set. To enhance these anchors, CAM retrieves analogous molecules and captures task-specific contextual knowledge from the unlabeled reference to obtain contextual representation anchors. 
AAM then leverages these anchors to transfer enriched contextual knowledge to individual molecules, enhancing their representations and alleviating sample selection bias. Finally, MM generates predictions for the query molecules.

\subsubsection{Context Augmentation Module}
\label{sss:context_calibrated_module}


The context augmentation module (CAM) enhances the representativeness of the initial anchors by utilizing unlabeled reference. In data-limited scenarios, the initial  anchors may not sufficiently represent their respective categories. CAM incorporates contextual information from the reference, enriching the class-level contextual representation anchors with task-specific knowledge and more accurately capturing the underlying molecular co-occurrence structures.

In Figure \ref{fig:method_pipeline}, given a new task that contains a support set $\mathcal{S}_{\tau}$ with an initial molecular feature matrix $\mathbf{S}_{\tau} \in \mathbb{R}^{ N_{\tau}^s \times d}$, a query set $\mathcal{Q}_{\tau}$ with an initial molecular feature matrix $\mathbf{Q}_{\tau} \in \mathbb{R}^{N_{\tau}^q \times d}$, and a batch of context $\mathcal{B}_{\tau}$ sampled from unlabeled data set $\mathcal{B}$ with an initial molecular feature matrix $\mathbf{B}_{\tau} \in \mathbb{R}^{M  \times d}$,
 we first employ a shared encoder $f_{\mathrm{e}}:\mathbb{R}^{d}\xrightarrow{}\mathbb{R}^{h}$ to obtain embeddings for all molecules as follows:
\begin{equation}\label{eq:1}
\mathbf{S}_{\tau}^\prime=f_{\mathrm{e}}(\mathbf{S}_{\tau}),\  \mathbf{Q}_{\tau}^\prime=f_{\mathrm{e}}(\mathbf{Q}_{\tau}),\  \mathbf{B}_{\tau}^\prime=f_{\mathrm{e}}(\mathbf{B}_{\tau}), 
\end{equation}
where $\mathbf{S}_{\tau}^\prime \in \mathbb{R}^{N_{\tau}^s \times h}$, $\mathbf{Q}_{\tau}^\prime \in \mathbb{R}^{N_{\tau}^q \times h}$, and $\mathbf{B}_{\tau}^\prime \in \mathbb{R}^{M \times h}$.

Next, we use the support molecular embedding matrix $\mathbf{S}_{\tau}^\prime$  to evaluate the initial class-level anchors: 
\begin{equation}\label{eq:2}
    \mathbf{p}_{\tau, c} = \frac{1}{N_{\tau, c}}\sum_{i=1}^{N_{\tau}}\mathbf{s}_{\tau, i}^\prime *\mathbbm{1}\{y_{\tau,i} = c\} , \ c \in \left\{-1, 1\right\},
\end{equation}
where $\mathbf{p}_{\tau, c} \in \mathbb{R}^{h}$ represents the embedding for class $c$, forming the initial class-level anchors $\mathbf{P}_{\tau}\in \mathbb{R}^{2 \times h}$. 

After acquiring the initial class-level anchors $\mathbf{P}_{\tau}$, CRA captures task-specific information from the contextual references and enhances the co-occurrence structures of these anchors.
Specifically, to fully leverage the benefits of rich contextual knowledge, we propose using the context $\mathcal{B}_{\tau}$ to enhance the few-shot data $\mathcal{T}_{\tau}$. 
Considering that molecules with similar structures have similar properties, we use the attention mechanism~\cite{vaswani2017attention} as a context retrieval mechanism to retrieve references with co-occurrence structural information, thereby enhancing the initial class-level anchors $\mathbf{P}_{\tau}$:
\begin{equation}\label{eq:3'}
    [\mathbf{P}_{\tau}^{\prime}:\mathbf{B}_{\tau}^{\ast}]_0 = \operatorname{R-MHA}\left([\mathbf{P}_{\tau}: \mathbf{B}_{\tau}^\prime]_0\right),
\end{equation}
where $\mathbf{P}_{\tau}^{\prime} \in \mathbb{R}^{2 \times h}$ represents the augmented contextual representation anchors, and $[\cdot : \cdot]_i$ denotes concatenating matrices in dimension $i$. $\mathbf{B}_{\tau}^{\ast}$ is an auxiliary output that is not used in subsequent steps.

Introducing the context $\mathcal{B}_{\tau}$ allows the augmented $\mathbf{P}_{\tau}^{\prime}$ to capture the core category features more comprehensively and exhibit higher representativeness. This prevents $\mathbf{P}_{\tau}^{\prime}$ from overfitting to specific minority molecules and generalizes it better to new molecules.

\subsubsection{Anchor Augmentation Module}
\label{sss:prototype-prompt_attention}
To enhance the intra-class representativeness of individual molecules for a given task $\mathcal{T}_{\tau}$, we adopt the contextual representation anchors $\mathbf{P}_{\tau}^{\prime}$ as bridges to evoke co-occurrence structures and semantic information within each class. 
Specifically, we concatenate each molecular embedding with  both anchors from the positive and negative classes to form the anchor-augmented embedding matrices $\mathbf{S}_{\tau}^{\prime\prime} \in \mathbb{R}^{N_{\tau}^s \times 3h}$ and $\mathbf{Q}_{\tau}^{\prime\prime} \in \mathbb{R}^{N_{\tau}^q \times 3h}$ as follows:
\begin{equation}\label{eq:concate}
    \mathbf{S}_\tau^{\prime\prime} = [\mathbf{S}_\tau^{\prime}, \mathbf{P}_{\tau,-1}^\prime, \mathbf{P}_{\tau, 1}^\prime]_{1}, 
     \mathbf{Q}_\tau^{\prime\prime} = [\mathbf{Q}_\tau^{\prime}, \mathbf{P}_{\tau,-1}^\prime, \mathbf{P}_{\tau, 1}^\prime]_{1},    
\end{equation}
and then use a multi-head attention module to achieve knowledge transfer and integration among them:
\begin{equation}\label{eq:4'}
    \left[[\mathbf{S}_{\tau}^{\ast}: \mathbf{Q}_{\tau}^{\ast}]_0 : \mathbf{P}^{(N_{\tau}^s+N_{\tau}^q)\times2h}\right]_1 = \operatorname{R-MHA}\left(\left[\mathbf{S}_{\tau}^{\prime\prime}: \mathbf{Q}_{\tau}^{\prime\prime}\right]_0\right), 
\end{equation}
where $\mathbf{S}_{\tau}^{\ast} \in \mathbb{R}^{ N_{\tau}^s \times h}$ and $\mathbf{Q}_{\tau}^{\ast} \in \mathbb{R}^{N_{\tau}^q \times h}$.

With rich class-level contextual information as prior knowledge, AAM infuses molecules with co-occurrence structures and semantic information, enhancing their representativeness. Consequently, the distance between molecules within the same category is reduced, while the distance between molecules of different categories is increased.

\subsubsection{Matching Module}
\label{sss:matching_module}
Following previous work~\cite{koch2015siamese, schimunek2023contextenriched}, we employ a cosine similarity function $sim(\cdot,\cdot)$ to measure the similarity between the query molecular embedding $\mathbf{q}_{\tau, j}^{\ast}$ and the support molecular embedding $\mathbf{s}_{\tau, i}^{\ast}$ within the task $\mathcal{T}_{\tau}$. Here, $\mathbf{q}_{\tau, j}^{\ast}$ represents the $j$-th embedding of $\mathbf{Q}_{\tau}^{\ast}$ and $\mathbf{s}_{\tau, i}^{\ast}$ represents the $i$-th embedding of $\mathbf{S}_{\tau}^{\ast}$. 
The similarity value $sim(\mathbf{q}_{\tau, j}^{\ast}, \mathbf{s}_{\tau, i}^{\ast})$ serves as the weight coefficient for the corresponding support molecular label $\mathbf{y}_{\tau, i} $. The prediction is then obtained by:
\begin{equation}\label{eq:simi_function'}
    p(\hat{y}_{\tau, j}=1)=\sigma\left(\dfrac{1}{\sqrt{2h}} \sum_{i=1}^{N_{\tau}^s} \dfrac{y_{\tau, i}}{N_{\tau}^s(y_{\tau, i})} sim\left(\mathbf{q}_{\tau, j}^{\ast}, \mathbf{s}_{\tau, i}^{\ast}\right)\right) , 
\end{equation}
where $\sigma(\cdot)$ is the sigmoid function, and $N_{\tau}^s(y_{\tau, i})$ is the number of molecules with label $y_{\tau, i}$ in the support set $\mathcal{S}_{\tau}$. Both $N_{\tau}^s(y_{\tau, i})$ and $\sqrt{2h}$ prevent the logit from being too large, ensuring stable training. 
Additionally, $N_{\tau}^s(y_{\tau, i})$ addresses the class imbalance issue present in the FS-Mol dataset~\cite{stanley2021fsmol}. The probability of $p(\hat{y}_{\tau, j}=-1)$ is given by $1-p(\hat{y}_{\tau, j}=1)$.

\subsection{The Role of the Attention Mechanism}
In our method, the challenge is to capture relationships between limited labeled data and unlabeled molecules, where the anchors and attention mechanism play crucial roles in leveraging contextual information to enhance model generalization. CRA utilizes multi-head attention to capture subtle yet critical structural similarities, enabling better generalization in few-shot learning tasks, as shown in Figure~\ref{fig:method_pipeline}. The attention mechanism serves two primary purposes:
\begin{itemize}
    \item \textbf{Contextual retrieval}: Attention helps retrieve relevant contextual knowledge from unlabeled reference molecules, thereby enriching the class-level anchors. In multi-head attention, the model learns a set of attention scores, (i.e., similarity measures)  by comparing the class-level anchors to the unlabeled reference molecules. Molecules deemed more relevant to the class are assigned higher attention scores, allowing the model to focus on those reference molecules. Then, the model aggregates the knowledge from these molecules to augment the anchors, making them more representative of the underlying class.  
    
    \item \textbf{Knowledge transfer}: 
    After augmenting the anchors, we concatenate each molecular representation with the augmented positive and negative anchors. This allows each molecular representation to reflect not only its own features but also the task's global context. In other words, this concatenation places the molecule in a broader context, enabling the model to better understand the relative relationships between the molecule and the corresponding anchors. Within this global task context, the attention mechanism captures task-specific similarities between molecular representations and dynamically adjusts the association strength between the molecules and the positive and negative anchors based on these similarities. This process helps the model focus on task-relevant molecular features while effectively filtering out irrelevant information, ensuring precise knowledge transfer and integration, and ultimately improving the model’s generalization ability in few-shot learning scenarios.
\end{itemize}

\subsection{Training and Inference}
We train our model using a set of training tasks $\left\{\mathcal{T}_{\tau}\right\}_{\tau=1}^{N_{t}}$. For each task $\mathcal{T}_\tau$, we sample two labeled sets, $\mathcal{S}_\tau, \mathcal{Q}_\tau$, from $\mathcal{T}_\tau$, and randomly sample an unlabeled reference set $\mathcal{B}_\tau$ from $\mathcal{B}$. The model is then trained to predict the labels in $\mathcal{Q}_\tau$ using $\mathcal{S}_\tau$ as the support set.
After obtaining the predicted probability $p(\hat{y}_{\tau, j}=c)$ for $c\in \{-1,1\}$ across all query molecules in the query set $\mathcal{Q}_{\tau}$, we use the Binary Cross Entropy (BCE) loss function to calculate the loss:
\begin{equation}\label{eq:7'}
    \mathcal{L} = -\frac{1}{N_{\tau}^q}\sum_{j=1}^{N_{\tau}^q}\sum_{c\in \{-1,1\}}\left[\mathbb{I}_c(y_{\tau, j})\cdot\log p(\hat{y}_{\tau, j}=c) \right],
\end{equation}
where $y_{\tau,j}$ is the ground-truth label of the $j$-th query molecule in $\mathcal{Q}_{\tau}$, and $\mathbb{I}_c(y_{\tau, j})$ is the indicator function that takes the value of $1$ if $y_{\tau, j}=c$, and $0$ otherwise.

Algorithm \ref{alg:testing} outlines the training procedure of \method. Line 3-8 correspond to the context augmentation process, which uses a wide range of reference molecules to augment the contextual representation anchors. Line 9 shows how these anchors act as bridges, transferring knowledge from the unlabeled reference data to the few-shot samples.
\begin{algorithm}
    \caption{Training procedure of \method}
    \label{alg:testing}
    \renewcommand{\algorithmicrequire}{\textbf{Input:}}
    \renewcommand{\algorithmicensure}{\textbf{Output:}}
    \begin{algorithmic}[1]
        \REQUIRE Molecular property prediction tasks $\left\{\mathcal{T}_{\tau}\right\}_{\tau=1}^{N_{t}}$, reference set $\mathcal{B}$, learning rate $\eta$, batch size of reference molecules $M$
        \ENSURE trained model $f_{\bm{\theta}}$
        \STATE Randomly initialize $\bm{\theta}$;
        \WHILE {not converged}
            \STATE Sample a  task $\mathcal{T}_{\tau}$;
            \STATE Sample $N_{\tau}^{s}$ and $N_{\tau}^{q}$ molecules to form support set $\mathcal{S}_{\tau}$ and query set $\mathcal{Q}_{\tau}$, respectively;
            \STATE Sample $M$ unlabeled reference molecules $B_{\tau}$;
            \STATE Obtain molecular representations $\mathbf{S}_{\tau}^{\prime}$, $\mathbf{Q}_{\tau}^{\prime}$ and $\mathbf{B}_{\tau}^{\prime}$ by Equation \ref{eq:1};
            \STATE Evaluate the initial class-level anchors $\mathbf{p}_{\tau, c}$ by Equation \ref{eq:2}; 
            \STATE Augment the initial class-level anchors $\mathbf{p}_{\tau,c}$ by Equation \ref{eq:3'} to obtain refined anchors $\mathbf{p}_{\tau,c}^\prime$;
            \STATE Concatenate each molecular embedding with all anchors to form anchor-augmented embeddings $\mathbf{S}_{\tau}^{\prime\prime}$ and $\mathbf{Q}_{\tau}^{\prime\prime}$ by Equation~\ref{eq:concate};
            \STATE Debias molecular representations $\mathbf{S}_{\tau}^{\prime}$ and $\mathbf{Q}_{\tau}^{\prime}$ by Equation \ref{eq:4'} to obtain refined molecular representations $\mathbf{S}_{\tau}^{\ast}$ and $\mathbf{Q}_{\tau}^{\ast}$;
            \STATE Obtain class prediction $\hat{p}(y_{\tau, j}=c)$ for all $c$ by Equation~\ref{eq:simi_function'};
            \STATE Calculate $\mathcal{L}$ by Equation \ref{eq:7'};
            \STATE Update $\bm{\theta} \leftarrow \bm{\theta} - \eta \nabla_{\bm{\theta}}\mathcal{L}$;
        \ENDWHILE
    \end{algorithmic}
  
\end{algorithm}

After the  training phase, the model is evaluated on a set of new test tasks $\left\{\mathcal{T}_{\tau}\right\}_{\tau=N_{t}+1}^{N_{t}+N_{e}}$ in a few-shot setting. For each task, we apply Equation~\ref{eq:simi_function'} to infer final prediction results. This evaluation process assesses the model's ability to generalize to unseen tasks by leveraging the knowledge learned during training, demonstrating its effectiveness in few-shot learning scenarios.

\section{Experiments}
In this section, we evaluate the performance of \textbf{CRA} as introduced in Section~\ref{subsection:CRA_network}. We first conduct experiments on the MoleculeNet and FS-Mol benchmarks to validate the efficacy of our method. 
To analyse the impact of various components on the model, we also conduct ablation experiments on the three components, the capacity of reference molecules, and the support set size during testing.
Additionally, we visualize the debiasing capability of CRA and the relationship between molecular similarities and attention weights. 
We conduct a domain shift experiment to validate the generalization of CRA. 
Finally, we evaluate the generalizability of CRA across different context sets.


\subsection{Few-shot Drug Discovery on MoleculeNet Benchmark}
\label{exp:moleculenet_benchmark}

\begin{table*}
 \centering
  \caption{All methods are compared on the MoleculeNet benchmark with a support set size of 20, with the mean test performance measured by AUROC\% and the corresponding standard deviations.
 }
  \label{Table:moleculenet}
  \begin{tabular*}{0.75\textwidth}{@{\extracolsep{\fill}}lcccccc@{\extracolsep{\fill}}}
    \toprule
     Method  & Tox21 [12] $\uparrow$   & SIDER [27] $\uparrow$   & MUV [17] $\uparrow$   & ToxCast [617] $\uparrow$ \\
    \midrule
    Siamese \cite{koch2015siamese}  & 80.40 $\pm$ 0.35  & 71.10 $\pm$ 4.32  & 59.59 $\pm$ 5.13  & - \\
    ProtoNet \cite{snell2017prototypical} & 74.98 $\pm$ 0.32  &  64.54 $\pm$ 0.89  & 65.88 $\pm$ 4.11  & 63.70 $\pm$ 1.26 \\
    MAML \cite{finn2017model}    & 80.21 $\pm$ 0.24   & 70.43 $\pm$ 0.76  & 63.90 $\pm$ 2.28 & 66.79 $\pm$ 0.85  \\
    TPN \cite{liu2018learning}     & 76.05 $\pm$ 0.24   & 67.84 $\pm$ 0.95  & 65.22 $\pm$ 5.82 & 62.74 $\pm$ 1.45 \\
    EGNN \cite{kim2019edge}  & 81.21 $\pm$ 0.16   & 72.87 $\pm$ 0.73  & 65.20 $\pm$ 2.08 & 63.65 $\pm$ 1.57 \\ 
    IterRefLSTM \cite{altae2017low} & 81.10 $\pm$ 0.17 & 69.63 $\pm$ 0.31 & 45.56 $\pm$ 5.12  & - \\ 
    PAR \cite{wang2021propertyaware} & 82.06 $\pm$ 0.12   &   \textbf{74.68} $\pm$ \textbf{0.31}   &   66.48 $\pm$ 2.12   & 69.72 $\pm$ 1.63 \\
    ADKF-IFT \cite{chen2023metalearning} & 82.43 $\pm$ 0.60 & 67.72 $\pm$ 1.21  & \textbf{98.18} $\pm$ \textbf{3.05} & 72.07 $\pm$ 0.81 \\ 
    MHNFs \cite{schimunek2023contextenriched} &   80.23 $\pm$ 0.84    &  65.89 $\pm$ 1.17    & 73.81 $\pm$ 2.53  &
    74.91 $\pm$ 0.73 \\ 
    \textbf{CRA (Ours)} & \textbf{82.51} $\pm$ \textbf{0.41} & 72.12 $\pm$ 1.23 & 77.31 $\pm$ 1.78 & \textbf{77.32} $\pm$ \textbf{0.64} \\
    \midrule
    Pre-GNN \cite{hu2019strategies} & 82.14 $\pm$ 0.08  & 73.96 $\pm$ 0.08  & 67.14 $\pm$ 1.58 & 73.68 $\pm$ 0.74 \\ 
    Meta-MGNN \cite{guo2021few} & 82.97 $\pm$ 0.10 & 75.43 $\pm$ 0.21 & 68.99 $\pm$ 1.84 & - \\ 
    Pre-PAR \cite{wang2021propertyaware} & 84.93 $\pm$ 0.11 & 78.08 $\pm$ 0.16 & 69.96 $\pm$ 1.37 & 75.12 $\pm$ 0.84 \\ 
    \textbf{Pre-CRA (Ours)} & \textbf{86.41} $\pm$ \textbf{0.39} & \textbf{80.23} $\pm$ \textbf{0.75} & 80.43 $\pm$ 0.34 & \textbf{79.24} $\pm$ \textbf{0.91} \\
    \bottomrule
  \end{tabular*}
\end{table*}

\paragraph{Benchmark and Baselines}
The MoleculeNet benchmark \cite{wu2018moleculenet} is designed for the few-shot molecular property prediction task. It includes four small molecular datasets: Tox21 \cite{richard2020tox21}, SIDER \cite{kuhn2016sider}, MUV \cite{rohrer2009maximum}, and ToxCast \cite{richard2016toxcast}. These datasets cover drug toxicity, side effects, and bioactivity testing, making them popular in drug discovery. We select representative baselines and divide them into two types depending on whether the model is trained from scratch: 
1) Methods that train models from scratch: 
\begin{itemize}
    \item  \textbf{Siamese} \cite{koch2015siamese} utilizes dual conventional neural networks to  quantify the similarity between input molecule pairs. 
    \item  \textbf{ProtoNet} \cite{snell2017prototypical} employs instances from each category to compute class prototypes and assigns each query molecule to the closest prototypes based on their similarity.
    \item  \textbf{MAML} \cite{finn2017model} utilizes a meta-learning strategy for model training and adapts the meta-learned parameters to handle new tasks. 
    \item  \textbf{TPN} \cite{liu2018learning} uses a relation graph to facilitate label propagation in the context of transductive learning.
    \item  \textbf{EGNN} \cite{kim2019edge} is based on an edge-labeling graph that facilitates explicit cluster evolution through iterative updates of edge labels. 
    \item  \textbf{IterRefLSTM} \cite{altae2017low}  enhances the ability to learn distance metrics for small molecules by integrating residual LSTM embedding with graph convolutional neural networks.  
    \item  \textbf{PAR} \cite{wang2021propertyaware} leverages prototypes to obtain property-aware representations and implements label propagation using a relation graph. 
    \item  \textbf{MHNfs} \cite{schimunek2023contextenriched} utilizes a large number of references to enhance molecular representations in few-shot tasks and then leverages 
    an attention mechanism to fuse the enriched representations.  
    \item  \textbf{ADKF-IFT} \cite{chen2023metalearning} utilizes deep kernel Gaussian Processes (GPs) for adaptive deep kernel fitting, which is particularly beneficial for addressing challenges in drug discovery. 
\end{itemize}
 2) Methods that fine-tune pretrained model: 
 \begin{itemize}
     \item  \textbf{Pre-GNN} \cite{hu2019strategies} utilizes self-supervised learning techniques, such as the masked attribute task, to pre-train a GNN model, enabling it to learn more robust representations. 
     \item  \textbf{Meta-MGNN} \cite{guo2021few} is a meta-learning-based GNN designed to improve generalization in molecular property prediction tasks by learning to adapt to new tasks with limited data. 
     \item  \textbf{Pre-PAR} \cite{wang2021propertyaware} leverages a pre-trained GNN model as the backbone to enhance PAR.  
 \end{itemize}

 Our proposed methods, \textbf{CRA} and \textbf{Pre-CRA}, belong to the first and second types, respectively.
  All methods utilize GIN \cite{xu2018how} as the graph encoder, with pre-trained parameters provided by Pre-GNN \cite{hu2019strategies}. 

\paragraph{Evaluation Procedure}
In the MoleculeNet benchmark, we follow the experimental setup of PAR \cite{wang2021propertyaware}, maintaining the same configuration for the few-shot task, with a support set size of 20 (i.e., 2-way 10-shot) and a query set size of 16. 
The area under the receiver operating characteristic curve (AUROC) is used as the evaluation metric, and Adam \cite{kingma2014adam} is used as the optimizer to train our model. 
All results are based on  10 repeated experiments with different random seeds. 

\paragraph{Performance} 
Tabel~\ref{Table:moleculenet} shows the comparison of results between our \textbf{CRA} and \textbf{Pre-CRA} methods and all baselines. Our methods achieve the best performance on the Tox21, SIDER (pre-trained version), and ToxCast datasets. Specifically, CRA and Pre-CRA surpass the state-of-the-art (SOTA) by 0.10\% and 0.41\%, respectively, on the Tox21 dataset. They outperform the SOTA by 13.08\% on the SIDER dataset and by 3.22\% and 3.96\% on the ToxCast dataset. Additionally, our Pre-CRA ranks the second in performance on the MUV dataset.
Overall, the results of our methods demonstrate that contextual representation anchors which serve as bridges, transferring co-occurrence knowledge from unlabeled data to intra-class individual molecules, enhance model performance.

\subsection{Few-shot Drug Discovery on FS-Mol Benchmark}
\label{exp:fs_mol_benchmark}

\paragraph{Benchmark and Baselines}
FS-Mol, a molecular dataset designed for few-shot learning, primarily focuses on unbalanced tasks related to measuring various protein target activities \cite{stanley2021fsmol}. It consists of 233,786 compounds, 489,133 measurements, and 5,120 tasks. 
Following MHNfs \cite{schimunek2023contextenriched}, we compare \method with multiple baselines on the FS-Mol benchmark. 
Different from the MoleculeNet Benchmark, here we divide baselines into two types subject to whether transductive learning and semi-supervised supervised learning are conducted:
1) Method that do not use these strategies. On top of ProtoNet \cite{snell2017prototypical}, IterRefLSTM \cite{altae2017low}, and ADKF-IFT \cite{chen2023metalearning} previously mentioned, we select: 
\begin{itemize}
    \item \textbf{GNN-ST} \cite{stanley2021fsmol} utilizes message-passing neural networks (MPNNs) to enhance learning from graph-based molecular representations, particularly in quantum chemistry.
    \item \textbf{Molecule Attention Transformer} (\textbf{MAT}) \cite{maziarka2020molecule} adapts the Transformer architecture \cite{vaswani2017attention} by incorporating inter-atomic distances and molecular graph structures into the self-attention mechanism. 
    \item \textbf{Random Forest (RF)} \cite{breiman2001random}, as introduced by \cite{fabris2018new}, is a powerful ensemble method that enhances  predictive accuracy and model robustness by aggregating the decisions of multiple decision trees.
    \item \textbf{Multi-task GNN (GNN-MT)} \cite{stanley2021fsmol} employs a 10-layer pre-trained GNN combined with a meta-learning strategy to capture task-specific knowledge for various tasks. 
\end{itemize}
Methods that belong to transductive learning and semi-supervised learning. On top of PAR \cite{wang2021propertyaware} and MHNfs \cite{schimunek2023contextenriched}, we select:
\begin{itemize}
    \item \textbf{TPN} \cite{liu2018learning} utilizes a relation graph to carry out label propagation within a transductive learning framework.
    \item  \textbf{PRWN} \cite{ayyad2021semi} combines Prototype Networks with random walk to learn more comprehensive molecular representations. 
\end{itemize}


\paragraph{Evaluation Procedure}
Following FS-Mol \cite{stanley2021fsmol}, we adopt the $\Delta$AUC-PR metric to evaluate model performance. 
This metric is particularly useful in scenarios involving imbalanced datasets, as it provides a more accurate reflection of performance by focusing on the area under the precision-recall curve, which is crucial when class distributions are imbalanced. 
For each task, we employ imbalanced sampling with a fixed support set size of 16, while varying the ratio of positive to negative samples. 
This variation better simulates real-world conditions where imbalanced data is a common challenge. 
To ensure the robustness and generalizability of our CRA, we report all results as the mean of five independent training iterations, with each iteration utilizing ten different support set samples. This comprehensive evaluation protocol accounts for both the variability introduced by imbalanced sampling and the stochastic nature of training, providing a more reliable estimate of model performance.

\begin{table*}[htb]
\centering
\caption{Results of $\Delta$AUC-PR on FS-Mol. The best method is indicated in bold. Error bars show the standard errors across tasks in each category. All metrics are calculated over five training reruns and ten selections of support sets according to the method of~\cite{stanley2021fsmol}. The number of tasks in each category is noted in square brackets.} 

\begin{threeparttable}
\begin{tabular*}{0.7\textwidth}{@{\extracolsep{\fill}}lcccccc@{\extracolsep{\fill}}}
\toprule[1pt]
Method  & All [157] $\uparrow$ & Kin. [125] $\uparrow$ & Hydrol. [20] $\uparrow$ & Oxid. [7] $\uparrow$    
\\
\midrule
GNN-ST \cite{stanley2021fsmol}          & .029 $\pm$ .004& .027 $\pm$ .004& .040 $\pm$ .018& .020 $\pm$ .016    
\\  
MAT \cite{maziarka2020molecule}              & .052 $\pm$ .005& .043 $\pm$ .005& .095 $\pm$ .019& .062 $\pm$ .024  
\\
Random Forest \cite{breiman2001random}    & .092 $\pm$ .007& .081 $\pm$ .009& .158 $\pm$ .028& .080 $\pm$ .029 
\\
GNN-MT \cite{stanley2021fsmol}           & .093 $\pm$ .006& .093 $\pm$ .006& .108 $\pm$ .025& .053 $\pm$ .018
\\
PAR \cite{wang2021propertyaware}                 & .164 $\pm$ .008& .182 $\pm$ .009 & .109 $\pm$ .020& .039 $\pm$ .008
\\
MHNfs \cite{schimunek2023contextenriched}       & .194 $\pm$ .006 & .206 $\pm$ .012  & .167 $\pm$ .014& .114 $\pm$ .021
\\
TPN \cite{liu2018learning}                                             & .218 $\pm$ .005  & .226 $\pm$ .006 & .207 $\pm$ 0.24& .104 $\pm$ .007 
\\
ProtoNet \cite{snell2017prototypical}       & .212 $\pm$ .012& .226 $\pm$ .007& .188 $\pm$ .022 & .077 $\pm$ .031
\\
IterRefLSTM \cite{altae2017low}        & .234 $\pm$ .010 & .251 $\pm$ .010 & .199 $\pm$ .026 & .098 $\pm$ .027
\\
ADKF-IFT \cite{chen2023metalearning}         & .234 $\pm$ .009 & .248 $\pm$ .020& \textbf{.217 $\pm$ .017} & .106 $\pm$ .008
\\
PRWN \cite{ayyad2021semi}                                        & .236 $\pm$ .007 & .241 $\pm$ .009 & .184 $\pm$ .014 & .132 $\pm$ 0.12 
\\

\textbf{\method (Ours)} & \textbf{.244 $\pm$ .009} & \textbf{.259 $\pm$ .010} & .206 $\pm$ .013 & \textbf{.158 $\pm$ .017}
\\
\bottomrule[1pt]
\end{tabular*}
\begin{tablenotes}[para, flushleft]
    \footnotesize
\end{tablenotes}
\end{threeparttable} 
\label{fs_main_experiment}
\end{table*}

\paragraph{Performance}

Table~\ref{fs_main_experiment} presents a detailed comparison of the results between our proposed CRA and all baseline approaches with a support set size of 16. In this evaluation, which spans 157 test tasks, CRA demonstrates an average performance improvement of 3.28\% over the current state-of-the-art (SOTA) methods. This consistent improvement highlights the strength of CRA in addressing challenges related to sample selection bias, a critical issue in few-shot learning. Notably, CRA exhibits significant gains in the Kinematics (Kin.) and Oxidation (Oxid.) sub-task sets, outperforming SOTA methods by 3.19\% and an impressive 19.70\%, respectively. These results underscore the ability of CRA to excel in tasks with complex and varied molecular structures. However, it is important to note that in the Hydrology (Hydrol.) sub-task set, our method underperforms compared to the baselines, indicating potential areas for further refinement. 

\subsection{Ablation Study}
\label{sub:ablation study}

\paragraph{The Components of CRA}
\label{para:the_components_of_CRA}

To demonstrate that: 1) utilizing references (context) can enhance class-level contextual representation anchors by introducing task-specific contextual knowledge; and 2) these augmented these anchors can serves as bridges to transfer contextual knowledge to molecules, thereby reinforcing their representations and alleviating the selection bias, we conduct ablation experiments on the AM, CAM, and AAM components, based on the baseline (encoder only). The AM (attention module) is part of the AAM for comparison purposes. While the AM employs the same attention mechanism as the AAM, its embedding matrices are not augmented by these anchors.
The experiments on the AAM and CAM evaluate the effectiveness of
these anchors using Equation~\ref{eq:2} following the molecular encoder.

\begin{figure*}[htbp]
\centering\includegraphics[width=0.7\textwidth,height=0.22\textwidth]{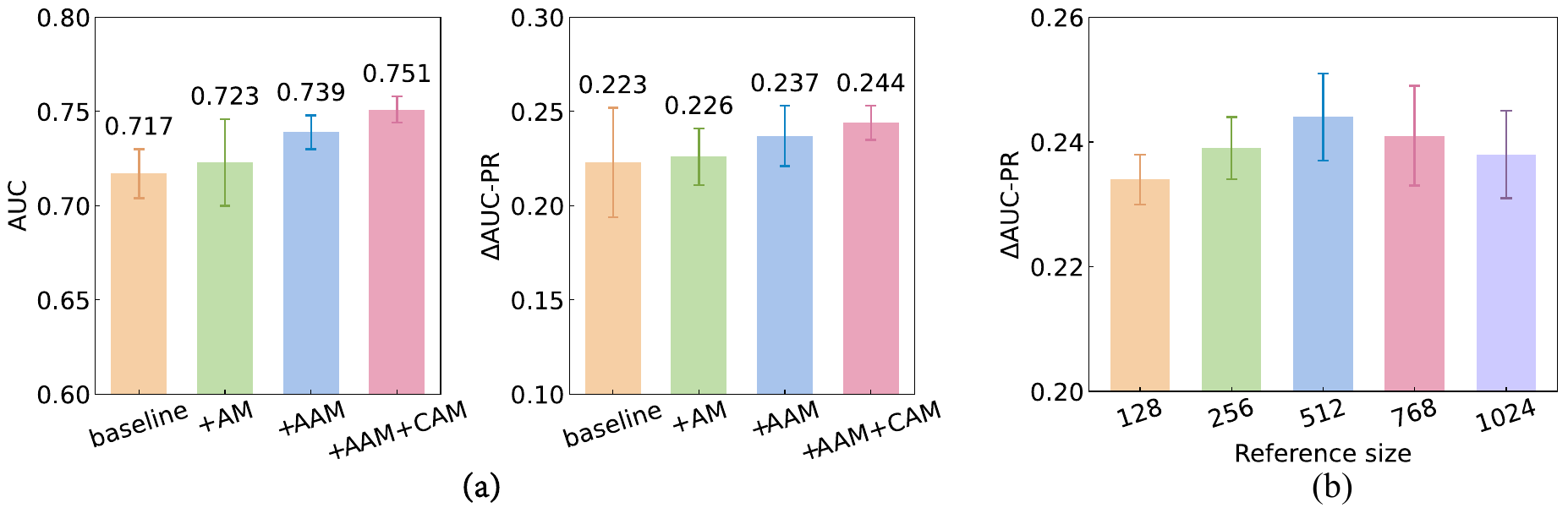}
    \vspace{-5pt}
    \caption{The mean performance with standard errors of the ablation study on FS-Mol benchmark. (a) The results (AUC and $\Delta$AUC-PR) of the ablation study for the components of CRA. (b) The performance of CRA with varying size of reference sets.}
    \label{fig:ablation_study}
\vspace{-20pt}
\end{figure*}

The results of the ablation study are shown in Figure~\ref{fig:ablation_study} (a), using AUC and $\Delta$AUC-PR as evaluation metrics. The Row labeled \textbf{``+ AM''} shows that the attention mechanism effectively captures relationships between molecules, as supported by prior work \cite{wang2021propertyaware, schimunek2023contextenriched}.
The results in Row \textbf{``+ AAM''} validate that using contextual representation anchors can transfer task-specific prior knowledge and enhance the expressiveness of molecular representations. Building on AAM, we further perform experiments on the CAM component. The results in Row \textbf{``+ AAM + CAM''} indicate that introducing unlabeled references can augment the representativeness of these anchors. These enriched anchors, serving as bridges, allow the task-specific contextual knowledge to be transferred to molecules, and thereby alleviate their bias.

\paragraph{Batch Size of Reference}
\label{para:batch_size_of_reference}
The component ablation experiments  have demonstrated the effectiveness of introducing references (context) to alleviate the sample selection bias. 
To further explore the benefits of using unlabeled reference data, we conduct additional ablation experiments by varying the reference set size. Figure~\ref{fig:ablation_study} (b) illustrates that
the performance of CRA initially improves with the increasing reference set size but then declines beyond a certain point. The optimal performance is achieved with a reference set containing 512 molecules.

\begin{figure}[htbp]
\centering\includegraphics[width=0.4\textwidth,height=0.25\textwidth]{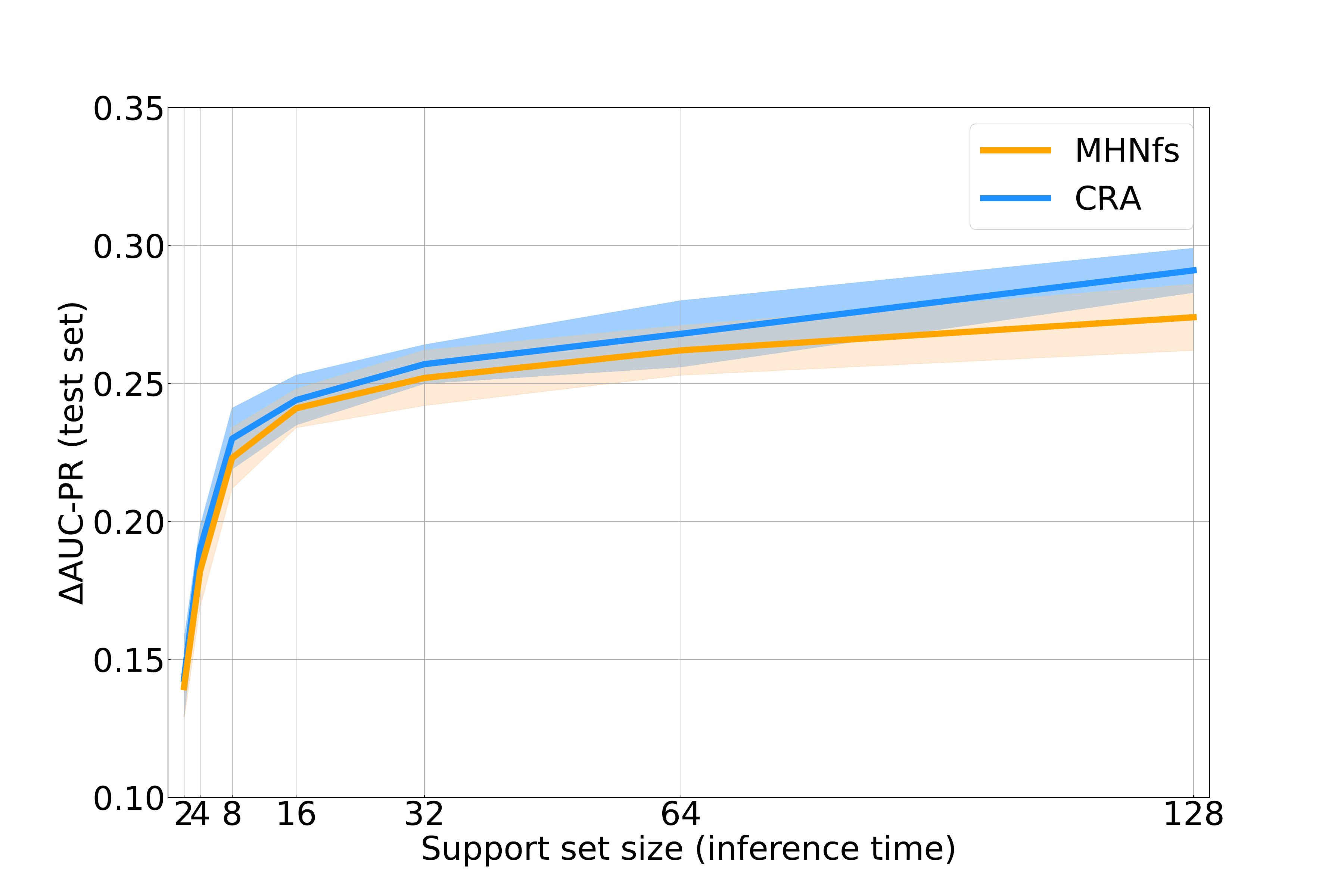}
    \caption{Performances of CRA and MHNfs for various support set sizes during inference.}
    \label{fig:support_set_study}
\end{figure}

\paragraph{Generalization to Different Support Set Sizes}
In addition, we assess the generalization ability of \method on new tasks with support sets of various sizes. As shown in Figure~\ref{fig:support_set_study}, following previous works~\cite{stanley2021fsmol, schimunek2023contextenriched}, we use support sets of sizes 2, 8, 16, 32, 64, and 128. The results show that, with smaller support sets, \method outperforms MHNfs \cite{schimunek2023contextenriched}, indicating that \method has considerable generalization ability for new tasks with limited support data. On larger support sets, the advantage of \method over MHNfs is even more significant, suggesting that both the introduction of a large amount of unlabeled data and the use of anchors to enhance the representativeness of molecular samples together effectively improve the model's generalization ability.

\begin{figure}[htbp]
\centering\includegraphics[width=0.45\textwidth,height=0.25\textwidth]{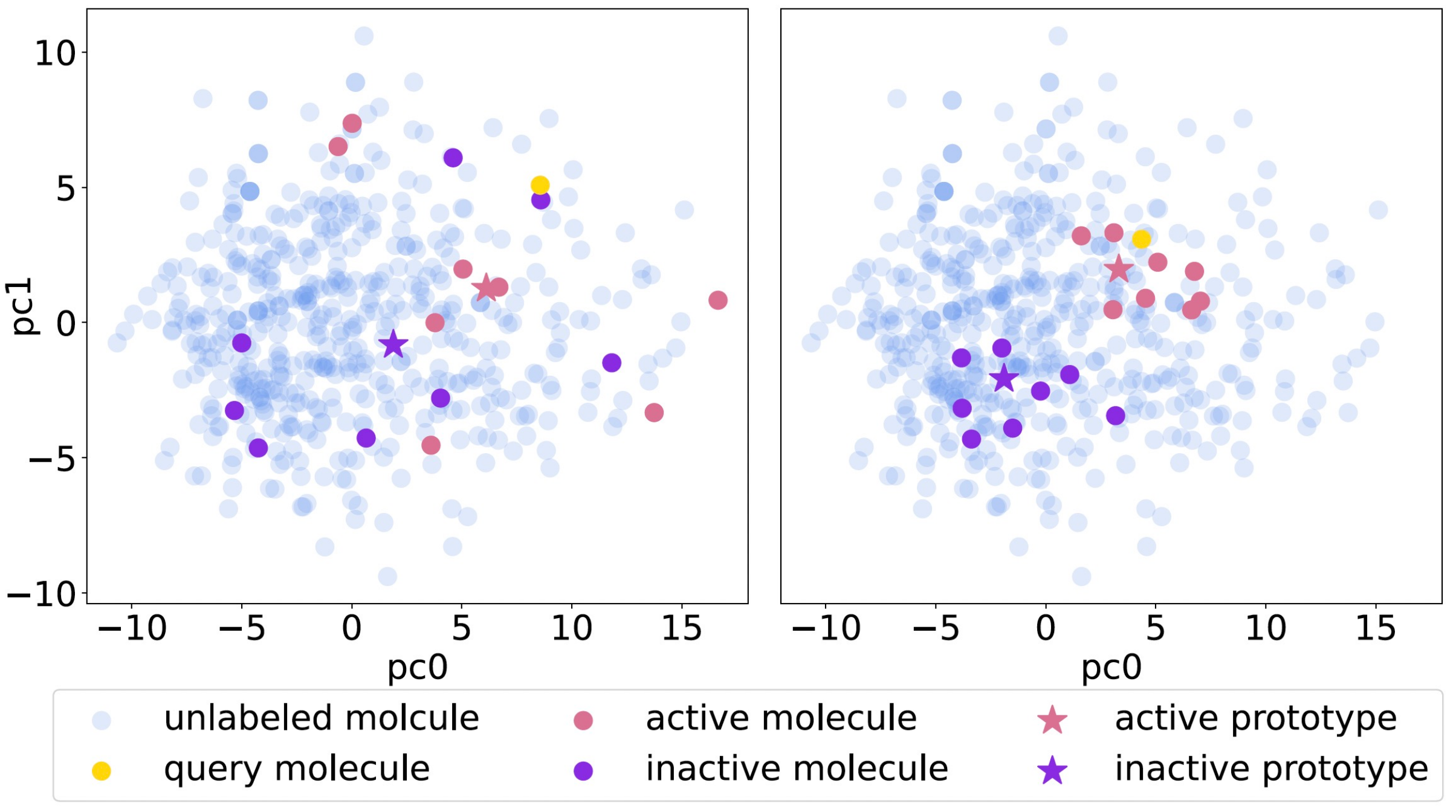}
    \caption{The visualization of \method on a few-shot task in chemical space. The task contains a query molecule and a support set with 8 molecules in each category. 512 unlabeled molecules are sampled as reference data. All molecular embeddings are compressed into a two-dimensional chemical space with PCA. The left chart displays the initial molecular embeddings and prototypes, while the right chart displays the enhanced molecular embeddings and calibrated prototypes. The initial embeddings of unlabeled molecules are shown in both charts for comparison.}
    \label{fig:visualization}
\end{figure}

\subsection{Visualization}
\paragraph{Visualization of CRA}
\label{para:visualization of cra}
To validate the efficacy of \method in enhancing molecular embeddings by introducing prior knowledge with calibrated class prototypes, we visualize the distribution of molecules in chemical space before and after enhancement. We randomly select a task from the test set with a support set of 16 molecules (8 positive and 8 negative) and 127 query molecules. From the 127 query molecules, we choose one for clearer visualization.
Additionally, we randomly extract 512 molecules from the unlabeled dataset as reference samples. Subsequently, we perform dimensionality reduction with PCA followed by visualization on both the initial molecular embeddings and the enhanced molecular embeddings along with the prototypes and calibrated prototypes. The visualization results are shown in Figure \ref{fig:visualization}. The left chart shows the visualization of the initial molecular embeddings, where positive and negative samples are intermixed, lacking clear classification boundaries. In contrast, the right chart shows the visualization of the enhanced molecular embeddings, with molecules of the same category clustered together and clear boundaries between positive and negative samples. These demonstrations demonstrate that by introducing context with rich structural information and mining the structural similarity  within the molecules in the task through calibrated class prototypes, we can significantly enhance the representativeness of molecular embeddings. This leads to the clustering of the same category and the separation of different categories. Therefore, positive and negative samples can be distinguished easily and the model's predictive performance is enhanced.

\paragraph{Visualization of Attention Matrix}

\begin{figure}
    \centering
    \includegraphics[width=0.98\linewidth]{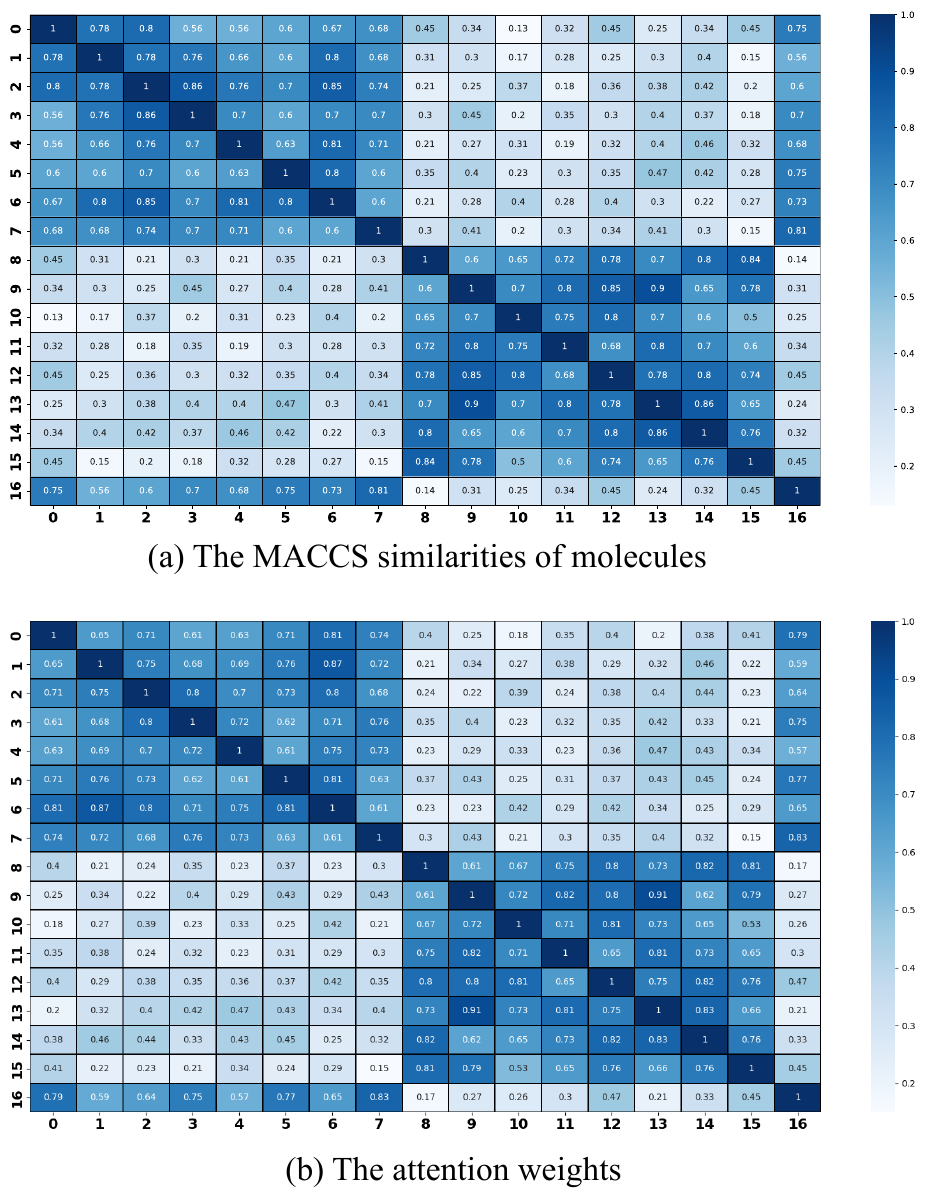}
    \caption{Heatmaps illustrating (a) the MACCS similarities among 16 molecules, with darker blue indicating higher similarity, and (b) the attention weights between molecule pairs, where darker shades represent stronger attention scores.}
    \label{fig:enter-label}
\end{figure}

From Figure \ref{fig:enter-label} (a) and (b), we can observe that the attention mechanism effectively captures the structural similarity between molecules. In Figure (a), the MACCS \cite{durant2002reoptimization} similarities provide a benchmark for the molecular similarity, while in Figure (b), the attention weights demonstrate that the model can assign stronger attention scores to molecule pairs with higher structural similarity. This suggests that the attention mechanism can serve as a powerful tool for identifying and emphasizing molecular relationships, enabling more accurate predictions in molecular property tasks.

\subsection{Domain Shift Experiment}
\label{sub:domain_shift_experiment}

\paragraph{Experiment Setup}
The Tox21 dataset is a public dataset for molecular toxicity, comprising 12 different toxicological tasks and 12,707 compounds \cite{mayr2016deeptox, richard2020tox21}. 
The data includes a variety of sources, such as drugs and chemicals from industrial and consumer products.
There is a significant domain shift from the FS-Mol dataset to the Tox21 dataset. The FS-Mol dataset considers the interactions between molecules and specific target proteins, while the Tox21 dataset 
is focus on the toxicities of small molecules across multiple proteins. 
To demonstrate the strong out-of-domain generalization ability of \method, we train three models---CRA, MHNfs \cite{schimunek2023contextenriched}, and PRWN \cite{ayyad2021semi}---on the FS-Mol dataset and evaluate their performances on the Tox21 dataset. 

\begin{table}[htb]
\centering
\caption{Results of domain shift experiments on the Tox21 dataset. The optimal method is in bold. Error bars represent the standard deviation across multiple training reruns and support set selections.}
\label{tab:domain shift experiment}
    \begin{tabular}{l|cccc}
    \toprule[1pt]
    Method           & AUC $\uparrow$   & $\triangle$ AUC-PR $\uparrow$ \\ \midrule
    MHNfs             & $.649\pm{.011}$ & $.087\pm{.009}$ &    \\
    PRWN       & $.675\pm{.009}$ & 
    $.094\pm{.014}$ \\
    \textbf{\method (Ours)}    & $\textbf{.700}\pm\textbf{.008}$ & $\textbf{.113}\pm\textbf{{.006}}$ &    \\ 
    \bottomrule[1pt]
    \end{tabular}
\vspace{-4pt}
\end{table}

\paragraph{Performance}
Table~\ref{tab:domain shift experiment} compares the generalization performance of MHNfs \cite{schimunek2023contextenriched}, PRWN \cite{ayyad2021semi}, and \method on the Tox21 dataset, and reports both AUC and $\Delta$AUC-PR. The evaluation metrics follow those of FS-Mol. Results show that \method achieves the best generalization on Tox21 dataset, with gains of 3.70\% in AUC and 20.20\% in $\Delta$AUC-PR.

\subsection{Generalization to Different Context Sets}
\label{sub:generalization to different context sets}
\begin{table}[]
    \centering
       \caption{CRA performance for different context sets $\Delta$AUC-PR. The error bars indicate the standard deviation across training reruns and different support set samples.}
    \begin{tabular}{l|c}
 
    \toprule[1pt]
         Dataset used as a context & $\Delta$AUC-PR \\  
          \midrule
         FS-Mol \cite{stanley2021fsmol} & .244 $\pm$ .009 \\
         GEOM \cite{axelrod2022geom} &   .245 $\pm$ .007 \\ 
    \bottomrule
    \end{tabular}
    \label{tab:context_sets}
\end{table}

To study the effect of different context sets on our model, we also conduct experiments using the GEOM dataset \cite{axelrod2022geom} as the context. The GEOM dataset is a standard dataset for studying molecular geometries, containing various molecules and their three-dimensional structural information. It is widely used in computational chemistry and molecular simulation research. We used 100,000 pre-processed molecules as the context. The experimental results are shown in Table~\ref{tab:context_sets}.
The results show that the choice of context dataset has a minimal impact on the performance of CRA, as both context sets yield nearly identical results. This indicates that CRA can effectively generalize across different context sets, with both yielding strong performance and minimal variation in outcomes.

\section{Conclusion}

In this work, we present the contextual representation anchor network (CRA) to address the challenge of sample selection bias in few-shot molecular property prediction. By implementing a dual augmentation mechanism—consisting of context augmentation and anchor augmentation—CRA dynamically incorporates task-specific contextual knowledge, enhancing the expressiveness of molecular representations. Our extensive evaluations on the MoleculeNet and FS-Mol benchmarks, along with domain transfer experiments on the Tox21 dataset, demonstrate that CRA consistently outperforms state-of-the-art methods in both AUC and $\Delta$AUC-PR metrics.
Furthermore, CRA’s ability to generalize across domains underscores its potential for robust applications in drug discovery, particularly where labeled data is limited. In future work, we aim to expand the application of CRA to broader molecular and biochemical datasets and to explore the integration of additional contextual information to enhance its performance further.

\section*{Acknowledgments}
This work is supported in part by National Key Research and Development Program of China 2022YFB4500300.



 
%

\newpage

\bibliography{bare_jrnl_new_sample4}
\bibliographystyle{IEEEtran}












\section{Biography Section}
 
\vspace{11pt}

\bf{If you include a photo:}\vspace{-33pt}
\begin{IEEEbiography}[{\includegraphics[width=1in,height=1.25in,clip,keepaspectratio]{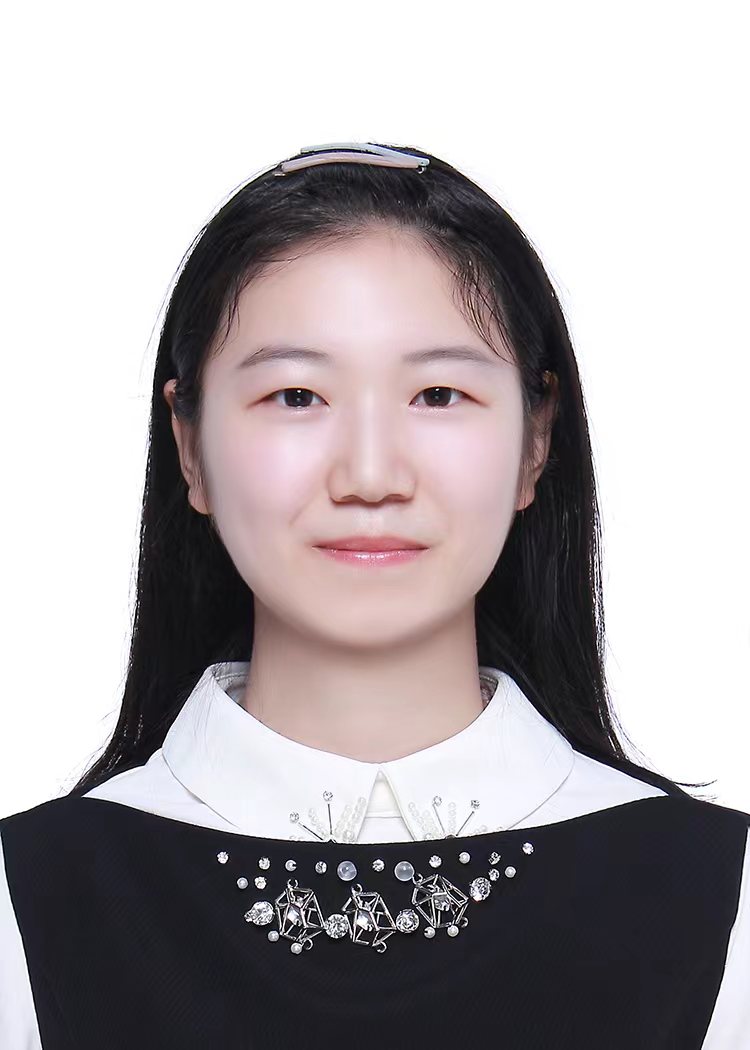}}]{Ruifeng Li}
is currently working toward the PhD degree with the College of Computer Science and Technology, Zhejiang University, jointly trained with the Zhejiang Lab. Her interest lies in AI-Driven Drug Design (AIDD) and Large Language Models (LLMs).  
\end{IEEEbiography}

\vspace{11pt}

\begin{IEEEbiography}[{\includegraphics[width=1in,height=1.25in,clip,keepaspectratio]{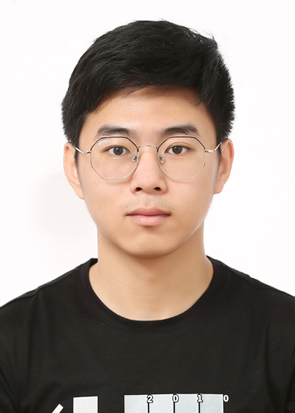}}]{Wei Liu} is currently a PhD student of the Department of Computer Science and Engineering, Shanghai Jiao Tong University, jointly trained with the Shanghai Artificial Intelligence Laboratory. In 2021, he got his bachelor's degree in Mechanical Engineering from Shanghai Jiao Tong University. His interest lies in AI-Driven Drug Design (AIDD) and Natural Language Processing (NLP). He has published works in conferences such as ICLR and NeurIPS.
\end{IEEEbiography}

\begin{IEEEbiography}[{\includegraphics[width=1in,height=1.25in,clip,keepaspectratio]{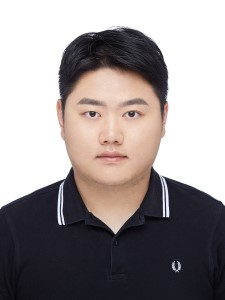}}]{Xiangxin Zhou} received the BS degree in electronic information science and technology from Tsinghua University, Beijing, China, in 2021, and he is currently working toward the PhD degree in School of Artificial Intelligence, University of Chinese Academy of Sciences and New Laboratory of Pattern Recognition (NLPR), State Key Laboratory of Multimodal Artificial Intelligence Systems (MAIS), Institute of Automation, Chinese Academy of Sciences (CASIA), Beijing, China. His research interests include generative models, geometric deep learning, and their application in AI for drug discovery.
\end{IEEEbiography}

\begin{IEEEbiography}[{\includegraphics[width=1in,height=1.25in,clip,keepaspectratio]{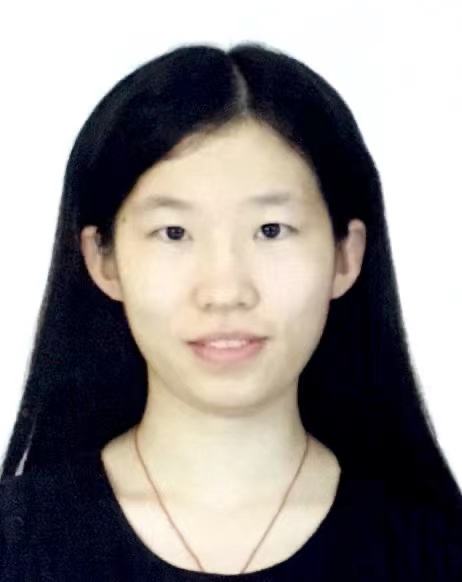}}]{Mingqian Li} currently works as a Post Doc in Zhejiang Lab, Hangzhou, China. She received the BS degree from the National University of Singapore, Singapore in 2016, and the PhD degree from Nanyang Technological University, Singapore in 2023. She has authored or co-authored papers in top tier conferences including AAAI, IJCAI, MobiCom, PAKDD. Her current research interests lie in graph data mining and spatio-temporal modeling.

\end{IEEEbiography}



\begin{IEEEbiography}
    [{\includegraphics[width=1in,height=1.25in,clip,keepaspectratio]{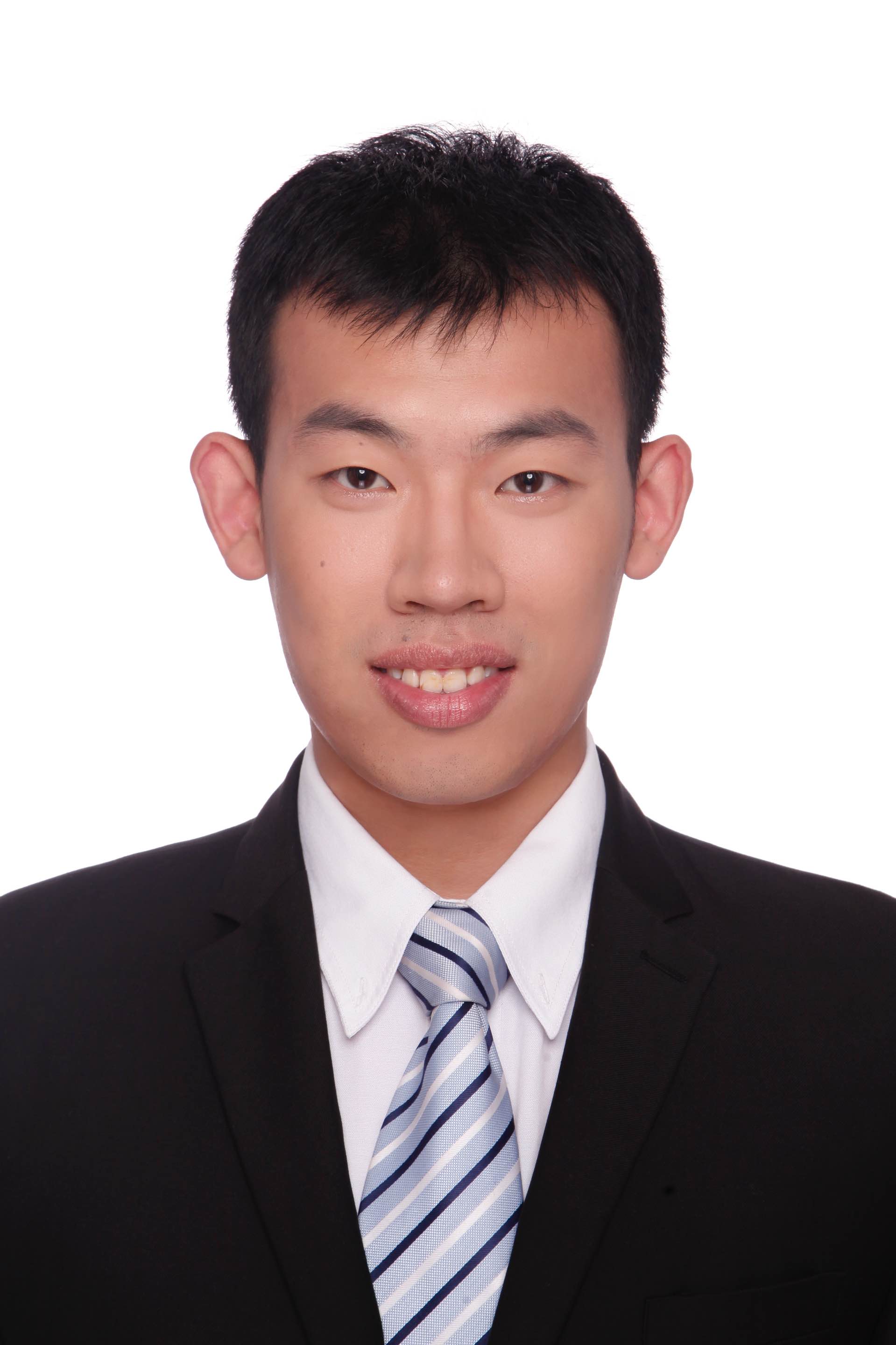}}]{Qiang Zhang} is an Assistant Professor at The Zhejiang University-University of Illinois Urbana-Champaign Institute (The ZJU-UIUC Institute), Zhejiang University in China. Before that, he obtained his Ph.D. degree and served as a postdoctoral researcher at the Department of Computer Science, University College London (UCL) in the UK. His research focuses on machine learning, natural language processing, knowledge graphs, and AI for Science. He has published over forty articles in top-tier AI academic journals and conferences including Nature Machine Intelligence, Nature Communications, NeurIPS, ICML, ICLR, AAAI and ACL. He also serves as the associate editor of the Big Data Research journal and PC member of top-tier AI conferences such as NeurIPS’19-24, ICML’19-24 and ICLR’18-24. 
\end{IEEEbiography}

\begin{IEEEbiography}
[{\includegraphics[width=1in,height=1.25in,clip,keepaspectratio]
{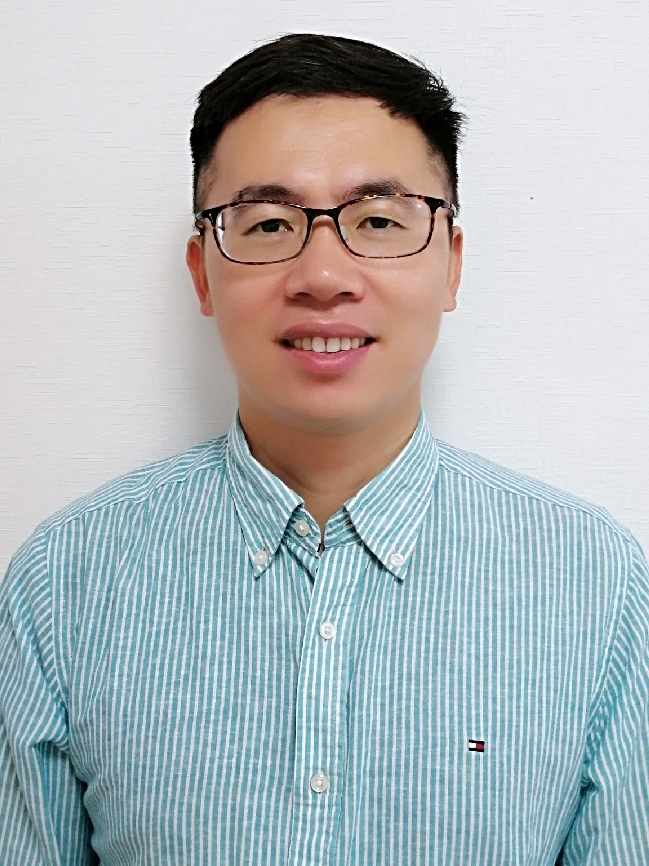}}]{Hongyang Chen} is the Deputy Director and Senior Research Expert (Senior Research Fellow) at the Data Hub and Security Research Center, Zhejiang Laboratory, and an IET Fellow. His research focuses on big data, artificial intelligence, and large-scale models. He has served as an editor for several well-known IEEE journals and as a field chair for IEEE international conferences. He has led national key research projects and National Natural Science Foundation of China (NSFC) projects. Dr. Chen has published over 100 papers in ACM/IEEE journals and CCF-A conferences. In the ICT field, he holds more than 30 international patents, several of which have been adopted as international standards. He received the Best Student Paper Award at ICDM 2023, was the global champion of the OGB Graph Machine Learning Challenge, and has been awarded the China Electronics Society Natural Science Award. In 2020, he was elected IEEE Distinguished Lecturer, and was recognized as the "2022 China Intelligent Computing Technology Innovation Figure" and as the 2023 "Young Pioneer of Computing Power in China."  
\end{IEEEbiography}

\begin{IEEEbiography}[{\includegraphics[width=1in,height=1.25in,clip,keepaspectratio]{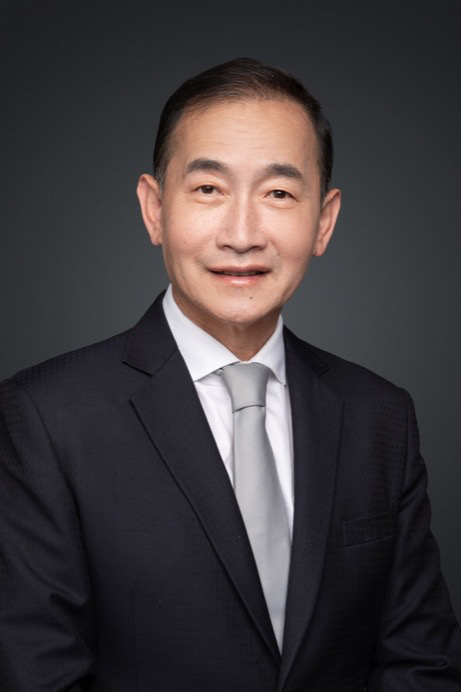}}]{Xuemin Lin}
    (Fellow, IEEE) received the BSc degree in applied math from Fudan University, in 1984, and the PhD degree in computer science from the University of Queensland, in 1992. He is a professor with Shanghai Jiao Tong University. His current research interests include the data streams, approximate query processing, spatial data analysis, and graph visualization.spatial data analysis, and graph visualization.
\end{IEEEbiography}



\vfill

\end{document}